\pdfoutput=1
    
\documentclass[11pt]{article}

\usepackage{acl}

\usepackage[shortlabels]{enumitem}

\usepackage{times}
\usepackage{latexsym}

\usepackage[T1]{fontenc}

\usepackage[utf8]{inputenc}

\usepackage{microtype}

\title{"That Is a Suspicious Reaction!": Interpreting Logits Variation to Detect NLP Adversarial Attacks}

\usepackage{tabularx}
\usepackage{colortbl}
\usepackage{graphicx}
\usepackage{subcaption}

\newcommand{\WDR}{\mathit{WDR}}

\author{Edoardo Mosca \\
  TU Munich,\\
  Department of Informatics,\\
  Germany\\
  \texttt{edoardo.mosca@tum.de} \\\And
  Shreyash Agarwal \\
  TU Munich,\\
  Department of Informatics,\\
  Germany\\
  \texttt{shreyash.agarwal@tum.de} \\\AND
  Javier Rando-Ramirez \\
  ETH Zurich, \\
  Department of Computer Science,\\
  Switzerland\\
  \texttt{jrando@student.ethz.ch} \\\And
  Georg Groh \\
  TU Munich,\\
  Department of Informatics,\\
  Germany\\
  \texttt{grohg@in.tum.de} \\
  }

\begin{document}
\maketitle
\begin{abstract}
Adversarial attacks are a major challenge faced by current machine learning research. These purposely crafted inputs fool even the most advanced models, precluding their deployment in safety-critical applications. Extensive research in computer vision has been carried to develop reliable defense strategies. However, the same issue remains less explored in natural language processing. Our work presents a model-agnostic detector of adversarial text examples. The approach identifies patterns in the logits of the target classifier when perturbing the input text. The proposed detector improves the current state-of-the-art performance in recognizing adversarial inputs and exhibits strong generalization capabilities across different NLP models, datasets, and word-level attacks.
\end{abstract}

\section{Introduction}

Despite recent advancements in \emph{Natural Language Processing} (NLP), adversarial text attacks continue to be highly effective at fooling models into making incorrect predictions \citep{ren2019generating, wang2019natural, garg2020bae}. In particular, syntactically and grammatically consistent attacks are a major challenge for current research as they do not alter the semantical information and are not detectable via spell checkers \citep{wang2019natural}. While some defense techniques addressing this issue can be found in the literature \citep{mozes2021frequency, zhou2019learning, wang2019natural}, results are still limited in performance and text attacks keep evolving. This naturally raises concerns around the safe and ethical deployment of NLP systems in real-world processes. 

Previous research showed that analyzing the model's logits leads to promising results in discriminating manipulated inputs \citep{wang2021modelagnostic, aigrain2019detecting, hendrycks2016early}. However, logits-based adversarial detectors have been only studied on computer vision applications. Our work transfers this type of methodology to the NLP domain and its contribution can be summarized as follows:
\paragraph{(1)} We introduce a logits-based metric called \emph{Word-level Differential Reaction} (WDR) capturing words with a suspiciously high impact on the classifier. The metric is model-agnostic and also independent from the number of output classes.
\paragraph{(2)} Based on WDR scores, we train an adversarial detector that is able to distinguish original from adversarial input texts preserving syntactical correctness. The approach substantially outperforms the current state of the art in NLP.
\paragraph{(3)} We show our detector to have full transferability capabilities and to generalize across multiple datasets, attacks, and target models without needing to retrain. Our test configurations include transformers and both contextual and genetic attacks.
\paragraph{(4)} By applying a post-hoc explainability method, we further validate our initial hypothesis---i.e. the detector identifies patterns in the WDR scores. Furthermore, only a few of such scores carry strong signals for adversarial detection.

\section{Background and Related Work} \label{sec:related_work}

\subsection{Adversarial Text Attacks}

Given an input sample $x$ and a target model $f$, an adversarial example $x' = x + \Delta x$ is generated by adding a perturbation $\Delta x$ to $x$ such that $\arg\max f(x) = y \neq y' = \arg\max f(x')$. Although this is not required by definition, in practice the perturbation $\Delta x$ is often imperceptible to humans and $x'$ is misclassified with high confidence. In the NLP field, $\Delta x$ consists in adding, removing, or replacing a set of words or characters in the original text. Unlike image attacks---vastly studied in the literature \citep{zhang2020adversarial} and operating in high-dimensional continuous input spaces---text perturbations need to be applied on a discrete input space. Therefore, gradient methods used for images such as FGSM \citep{goodfellow2014explaining} or BIM \citep{kurakin2016adversarial} are not useful since they require a continuous space to perturb $x$. Based on the text perturbation introduced, text attacks can be distinguished into two broad categories.

\paragraph{Visual similarity:} These NLP attacks generate adversarial samples $x'$ that look similar to their corresponding original $x$. These perturbations usually create typos by introducing perturbations at the character level. DeepWordBug \citep{gao2018blackbox}, HotFlip \citep{ebrahimi2018hotflip}
, and VIPER \citep{eger2019text} 
are well-known techniques belonging to this category.

\paragraph{Semantic similarity:} Attacks within this category create adversarial samples by designing sentences that are semantically coherent to the original input and also preserve syntactical correctness. Typical word-level perturbations are deletion, insertion, and replacement by synonyms \citep{ren2019generating} or paraphrases \citep{iyyer2018adversarial}. Two main types of adversarial search have been proposed. \emph{Greedy algorithms} try each potential replacement until there is a change in the prediction \citep{li2020bertattack,ren2019generating,jin2020bert}. On the other hand, \emph{genetic algorithms} such as \citet{alzantot2018generating} and \citet{wang2019natural} attempt to find the best replacements inspired by natural selection principles.

\subsection{Defense against Adversarial Attacks in NLP}

Defenses based on spell and syntax checkers are successful against character-level text attacks \citep{pruthi2019combating, wang2019natural, alshemali2019towards}. In contrast, these solutions are not effective against word-level attacks preserving language correctness \citep{wang2019natural}. We identify methods against word-level attacks belonging to two broad categories:

\paragraph{Robustness enhancement:} The targeted model is equipped with further processing steps to not be fooled by adversarial samples without identifying explicitly which samples are adversarial. For instance, \emph{Adversarial Training} (AT) \citep{goodfellow2014explaining} consists in training the target model also on manipulated inputs. The \emph{Synonym Encoding Method} (SEM) \citep{wang2019natural} introduces an encoder step before the target model's input layer and trains it to eliminate potential perturbations. Instead, \emph{Dirichlet Neighborhood Ensemble} (DNE) \citep{zhou2020defense} and \emph{Adversarial Sparse Convex Combination} (ASCC) \citep{dong2021towards} augment the training data by leveraging the convex hull spanned by a word and its synonyms.

\paragraph{Adversarial detection:} Attacks are explicitly recognized to alert the model and its developers. Adversarial detectors were first explored on image inputs via identifying patterns in their corresponding Shapley values \citep{fidel2020explainability}, activation of specific neurons \citep{tao2018attacks}, and saliency maps \citep{ye2020detection}. For text data, popular examples are \emph{Frequency-Guided Word Substitution} (FGWS) \citep{mozes2021frequency} and \emph{learning to DIScriminate Perturbation} (DISP) \citep{zhou2019learning}. The former exploits frequency properties of replaced words, while the latter uses a discriminator to find suspicious tokens and uses a contextual embedding estimator to restore the original word.

\begin{figure*}
    \centering
    \includegraphics[width=0.98\linewidth]{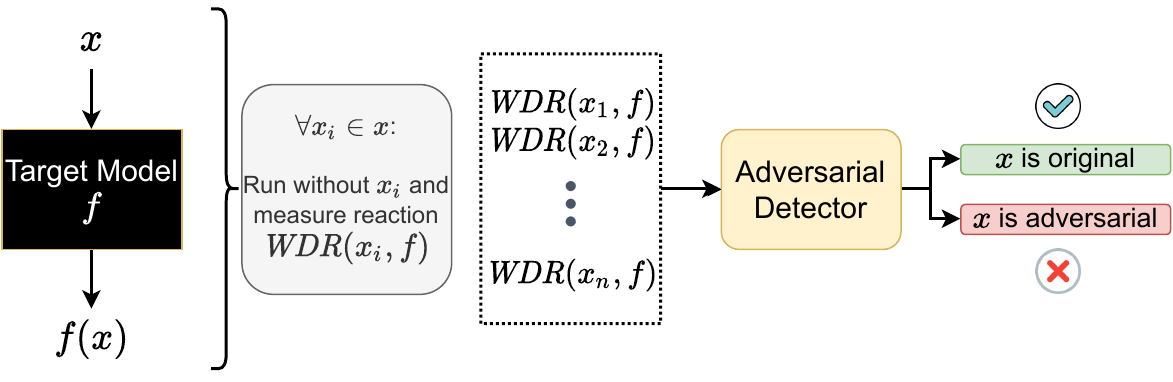}
    \caption{Overview of the proposed method.}
    \label{fig:pipeline}
\end{figure*}

\subsection{Logits-Based Adversarial Detectors}

Inspecting output logits has already led to promising results in discriminating between original and adversarial images \citep{hendrycks2016early, pang2018towards, kannan2018adversarial, roth2019odds}. For instance, \citet{wang2021modelagnostic} trains a recurrent neural network that captures the difference in the logits distribution of manipulated samples. \citet{aigrain2019detecting}, instead, achieves good detection performance by feeding a simple three-layer neural network directly with the logit activations. 

Our work adopts a similar methodology but focuses instead on the NLP domain and thus text attacks. In this case (1) logits-based metrics to identify adversarial samples should be tailored to the new type of input and (2) detectors should be tested on currently used NLP models such as transformers \citep{devlin-etal-2019-bert}.

\section{Methodology} \label{sec:methodology}

The defense approach proposed in this work belongs to the category of \emph{adversarial detection}. It defends the target model from attacks generated via word-level perturbations belonging to the \emph{semantic similarity} category. The intuition behind the method is that the model's reaction to original- and adversarial samples is going to differ even if the inputs are similar. Hence, it relies on \emph{feature attribution explanations} coupled with a machine learning model to learn such difference and thus identify artificially crafted inputs.

Figure \ref{fig:pipeline} shows the overall pipeline of the approach. Given a text classifier $f$ trained on the task at hand, the pipeline's goal is to detect whether the currently fed input $x$ is adversarial. In \ref{sec:logits}, we explain in greater detail how we measure the model $f$'s reaction to a given input $x$. This quantity---later indicated with $\WDR(x,f)$---is then passed to the adversarial detector, whose training procedure is described in \ref{sec:adv_training}. Finally, in \ref{sec:setup}, we provide detailed information about the setup of our experiments such as target models, datasets, and attacks.

\subsection{Interpreting the Target Model and Measuring its Reaction: Word-Level Differential Reaction} \label{sec:logits}

Adversarial attacks based on semantic similarity replace the smallest number of words possible to change the target model's prediction \cite{alzantot2018generating}. Thus, we expect the replacements transforming $x$ into $x'$ to play a big role for the output. If not, we would not have $f(x')$ substantially different from $f(x)$. To assess the reaction of the target model $f$ to a given input $x$, we measure the impact of a word via the \emph{Word-level Differential Reaction} (WDR) metric. Specifically, the effect of replacing a word $x_i$ on the prediction 

\[
y^* = \arg\max_{y} p(y| x)
\]

is quantified by

\[
\WDR(x_i,f) = f(x \backslash x_i)_{y^*} - \max_{y \neq y^*}f(x \backslash x_i)_y
\]

where  $f(x \backslash x_i)_y$ indicates the output logit for class $y$ for the input sample $x$ without the word $x_i$. Specifically, $x_i$ is replaced by an \emph{unknown word token}. If $x$ is adversarial, we could expect to find perturbed words to have a negative $\WDR(x_i, f)$ as without them the input text should recover its original prediction. Table \ref{tab:logits_example} shows an example pair of original and adversarial text together with their corresponding $\WDR(x_i, f)$ scores. The original class is recovered after removing a perturbed word in the adversarial sentence. This switch results in a negative WDR. However, even if the most important word is removed from the original sentence (\textit{'worst'}), the predicted class does not change and thus $\WDR(x_i, f) > 0$.

Our adversarial detector takes as input $\WDR(x, f)$, i.e. the sorted list of WDR scores $\WDR(x_i, f)$ for all words $x_i$ in the input sentence. As sentences vary in length, we pad the list with zeros to ensure a consistent input length for the detector.

\bgroup
\def\arraystretch{1.1}
\begin{table*}[!h]
\centering
\begin{tabularx}{\textwidth}{|l|l|l|l|ll|l|l|l|}
\cline{1-4} \cline{6-9}
\multicolumn{4}{|l|}{\textbf{\textcolor{ForestGreen}{Original sentence:} Neg. Review (\textit{Class 0})}}                                                                        & \multicolumn{1}{l|}{} & \multicolumn{4}{l|}{\textbf{\textcolor{Mahogany}{Adversarial sentence:} Pos. Review (\textit{Class 1})}}                                                                          \\ \cline{1-4} \cline{6-9} 
\multicolumn{4}{l}{This is absolutely   the worst trash I have ever}                                                          &                       & \multicolumn{4}{l}{This is absolutely the \textcolor{Mahogany}{tough} trash I have ever}                                                           \\
\multicolumn{4}{l}{seen. It took 15   full minutes before I realized}                                                         &                       & \multicolumn{4}{l}{seen. It took 15 full minutes before I realized}                                                          \\
\multicolumn{4}{l}{that what I was   seeing was a sick joke! [...]}                                                           &                       & \multicolumn{4}{l}{that what I was seeing was a \textcolor{Mahogany}{silly} joke! [...]}                                                           \\ \cline{1-4} \cline{6-9} 
\multicolumn{1}{|l|}{\textbf{Removed}}    & \multicolumn{1}{l|}{\textbf{Logit}} & 
\multicolumn{1}{l|}{\textbf{Logit}} & 
\multicolumn{1}{c|}{\textbf{WDR}} & \multicolumn{1}{l|}{} & \multicolumn{1}{l|}{\textbf{Removed}}    & \multicolumn{1}{l|}{\textbf{Logit}} & 
\multicolumn{1}{l|}{\textbf{Logit}} & \multicolumn{1}{c|}{\textbf{WDR}} \\
\multicolumn{1}{|c|}{\textbf{Word} $x_i$}    & \multicolumn{1}{l|}{\textbf{\textit{Class 0}}} & 
\multicolumn{1}{l|}{\textbf{\textit{Class 1}}} & \multicolumn{1}{c|}{$\WDR(x_i,f)$} & \multicolumn{1}{l|}{} & \multicolumn{1}{c|}{\textbf{Word} $x_i$}    & \multicolumn{1}{l|}{\textbf{\textit{Class 0}}} & \multicolumn{1}{l|}{\textbf{\textit{Class 1}}} & \multicolumn{1}{c|}{$\WDR(x_i,f)$} \\ \cline{1-4} \cline{6-9}
\multicolumn{1}{|l|}{$\emptyset$}  & \multicolumn{1}{l|}{3.44}    & \multicolumn{1}{l|}{-3.46}   & \multicolumn{1}{l|}{\textcolor{ForestGreen}{\textbf{6.89}}}    & \multicolumn{1}{l|}{} & \multicolumn{1}{l|}{$\emptyset$}  & \multicolumn{1}{l|}{-1.85}   & \multicolumn{1}{l|}{2.17}    & \multicolumn{1}{l|}{\textcolor{Mahogany}{\textbf{4.02}}}    \\ \cline{1-4} \cline{6-9} 
\multicolumn{1}{|l|}{worst}      & \multicolumn{1}{l|}{1.68}    & \multicolumn{1}{l|}{-1.75}   & \multicolumn{1}{l|}{\textcolor{ForestGreen}{\textbf{3.43}}}    & \multicolumn{1}{l|}{} & \multicolumn{1}{l|}{\textcolor{Mahogany}{tough}}      & \multicolumn{1}{l|}{2.14}    & \multicolumn{1}{l|}{-1.50}   & \multicolumn{1}{l|}{\textcolor{Mahogany}{\textbf{-3.64}}}   \\ \cline{1-4} \cline{6-9} 
\multicolumn{1}{|l|}{sick}       & \multicolumn{1}{l|}{3.34}    & \multicolumn{1}{l|}{-3.42}   & \multicolumn{1}{l|}{\textcolor{ForestGreen}{\textbf{6.76}}}    & \multicolumn{1}{l|}{} & \multicolumn{1}{l|}{\textcolor{Mahogany}{silly}}      & \multicolumn{1}{l|}{1.38}    & \multicolumn{1}{l|}{-1.37}   & \multicolumn{1}{l|}{\textcolor{Mahogany}{\textbf{-2.75}}}   \\ \cline{1-4} \cline{6-9} 
\multicolumn{1}{|l|}{absolutely} & \multicolumn{1}{l|}{3.40}    & \multicolumn{1}{l|}{-3.45}   & \multicolumn{1}{l|}{\textcolor{ForestGreen}{\textbf{6.86}}}    & \multicolumn{1}{l|}{} & \multicolumn{1}{l|}{absolutely} & \multicolumn{1}{l|}{-0.31}   & \multicolumn{1}{l|}{0.48}    & \multicolumn{1}{l|}{\textcolor{Mahogany}{\textbf{0.79}}}    \\ \cline{1-4} \cline{6-9} 
\multicolumn{1}{|l|}{realized}   & \multicolumn{1}{l|}{3.41}    & \multicolumn{1}{l|}{-3.47}   & \multicolumn{1}{l|}{\textcolor{ForestGreen}{\textbf{6.89}}}    & \multicolumn{1}{l|}{} & \multicolumn{1}{l|}{realized}   & \multicolumn{1}{l|}{-1.07}   & \multicolumn{1}{l|}{1.36}    & \multicolumn{1}{l|}{\textcolor{Mahogany}{\textbf{2.43}}}    \\ \cline{1-4} \cline{6-9}
\end{tabularx}
\caption{
$\WDR(x_i,f)$ scores computed for an \textcolor{ForestGreen}{original} sentence and its corresponding \textcolor{Mahogany}{adversarial} perturbation. Results show how when removing adversarial words such as \textit{tough} or \textit{silly}, the original class is recovered and the WDR becomes negative. $\emptyset$ corresponds to the prediction without any replacements}
\label{tab:logits_example}
\end{table*}
\egroup
\subsection{Adversarial Detector Training} \label{sec:adv_training}

The adversarial detector is a machine-learning classifier that takes the model's reaction $\WDR(x, f)$ as input and outputs whether the input $x$ is adversarial or not. To train the model, we adopt the following multi-step procedure:

\paragraph{(S1) Generation of adversarial samples:} Given a target classifier $f$, for each original sample available $x$, we generate one adversarial example $x'$. This leads to a balanced dataset containing both normal and perturbed samples. The labels used are \emph{original} and \emph{adversarial} respectively.
\paragraph{(S2) WDR computation:} For each element of the mixed dataset, we compute the $\WDR(x, f)$ scores as defined in Section \ref{sec:logits}. Once more, this step creates a balanced dataset containing the WDR scores for both normal and adversarial samples.
\paragraph{(S3) Detector training:} The output of the second step \textbf{(S2)} is split into training and test data. Then, the training data is fed to the detector for training along with the labels defined in step \textbf{(S1)}.
\paragraph{}

Please note that no assumption on $f$ is made. At the same time, the input of the adversarial detector---i.e. the WDR scores---does not depend on the number of output classes of the task at hand. Hence, the adversarial detector is model-agnostic w.r.t. the classification task and the classifier targeted by the attacks.

In our case, we do not pick any particular architecture for the adversarial detector. Instead, we experiment with a variety of models to test their suitability for the task. In the same spirit, we test our setting on different target classifiers, types of attacks, and datasets. 

\subsection{Experimental Setup} \label{sec:setup}

To test our pipeline, four popular classification benchmarks were used: \emph{IMDb} \citep{maas2011learning}, \emph{Rotten Tomatoes Movie Reviews} (RTMR) \citep{pang_2005_seeing}, \emph{Yelp Polarity} (YELP) \citep{zhang_2015_charachter}, and \emph{AG News} \citep{zhang_2015_charachter}. The first three are binary sentiment analysis tasks in which reviews are classified in either \emph{positive} or \emph{negative} sentiment. The last one, instead, is a classification task where news articles should be identified as one of four possible topics: \emph{World}, \emph{Sports}, \emph{Business}, and \emph{Sci/Tech}.

As main target model for the various tasks we use DistilBERT \citep{sanh2020distilbert} fine-tuned on IMDb. We choose DistilBert---a transformer language model \citep{vaswani2017attention}---as transformer architectures are widely used in NLP applications, established as state of the art in several tasks, and generally quite resilient to adversarial attacks \citep{morris2020textattack}. Furthermore, we employ a \emph{Convolutional Neural Network} (CNN) \citep{zhang_2015_charachter}, a \emph{Long Short-Term Memory} (LSTM) \citep{hochreiter_1997_long}, and a full BERT model \citep{devlin-etal-2019-bert} to test transferability to different target architectures. All models are provided by the TextAttack library \citep{morris2020textattack} and are already trained\footnote{textattack.readthedocs.io/en/latest/3recipes/models.html, released under MIT License} on the datasets used in the experiments.

We generate adversarial text attacks via four well-established word-substitution-based techniques: \emph{Probability Weighted Word Saliency} (PWWS) \citep{ren2019generating}, 
\emph{Improved Genetic Algorithm} (IGA) \citep{jia2019certified}, \emph{TextFooler} \citep{jin2020bert}, and \emph{BERT-based Adversarial Examples} (BAE) \citep{garg2020bae}. The first is a greedy algorithm that uses word saliency and prediction probability to determine the word replacement order \citep{ren2019generating}. 
IGA, instead, crafts attacks via mutating sentences and promoting the new ones that are more likely to cause a change in the output. TextFooler ranks words by importance and then replaces the ones with the highest ranks. Finally, BAE, leverages a BERT language model to replace tokens based on their context \citep{garg2020bae}. All attacks are generated using the TextAttack library \citep{morris2020textattack}.

We investigate several combinations of datasets, target models, and attacks to test our detector in a variety of configurations. Because of its robustness and well-balanced behavior, we pick the average F1-score as our main metric for detection. However, as in adversarial detection false negatives can have major consequences, we also report the recall on adversarial sentences. Later on, in \ref{subsec:boundary}, we also compare performance with other metrics such as precision and original recall and observe how they are influenced by the chosen decision threshold. 

\section{Experimental Results} \label{sec:results}

In this section, we report the experimental results of our work. In \ref{subsec:architecture_comparison}, we study various detector architectures to choose the best performing one for the remaining experiments. In \ref{subsec:performance}, we measure our pipeline's performance in several configurations (target model, dataset, attack) and we compare it to the current state-of-the-art adversarial detectors. While doing so, we also assess transferability via observing the variation in performance when changing the dataset, the target model, and the attack source without retraining our detector. Finally, in \ref{subsec:boundary}, we look at how different decision boundaries affect performance metrics.

\subsection{Choosing a Detector Model} \label{subsec:architecture_comparison}

The proposed method does not impose any constraint on which detector architecture should be used. For this reason, no particular model has been specified in this work so far. We study six different detector architectures in one common setting. We do so in order to pick one to be utilized in the rest of the experiments. Specifically, we compare XGBoost \citep{chen2016xgboost}, AdaBoost \citep{schapire1999brief}, LightGBM \citep{guolin2017lightgbm}, SVM \citep{hearst1998support}, Random Forest \citep{Breiman2001random}, and a Perceptron NN \citep{singh2019study}. All models are compared on adversarial attacks generated with PWWS from IMDb samples and targeting a DistilBERT model fine-tuned on IMDb. A balanced set of $3,000$ instances---$1,500$ normal and $1,500$ adversarial---was used for training the detectors while the test set contains a total of 1360 samples following the same proportions.

\begin{table}[!h] 
\centering
\begin{tabularx}{\linewidth}{|X|X|X|}
\hline
\textbf{Model} & \textbf{F1-Score} & \textbf{Adv. Recall}\\
\hline
\textbf{XGBoost} & \textbf{92.4} & 95.2 \\
\hline
AdaBoost & 91.8 & \textbf{96.0} \\
\hline
LightGBM & 92.0 & 93.7 \\
\hline
SVM & 92.0 & 94.8 \\
\hline
{\normalsize Random Forest} & 91.5 & 93.7 \\
\hline
Perceptron NN & 90.4 & 88.1 \\
\hline
\end{tabularx}
\caption{Performance comparison of different detector architectures on IMDb adversarial attacks generated with PWWS and targeting a DistilBERT transformer.} 
\label{tab:detector}
\end{table}

\begin{table*}[!h] 
\centering
\begin{subtable}{\textwidth}
\begin{tabularx}{\textwidth}{|X|X|X|X|X|X|X|}
\hline
\multicolumn{3}{|c|}{\textit{\textbf{Configuration}}} & \multicolumn{2}{c}{\textit{\textbf{WDR}} (Ours)} & \multicolumn{2}{|c|}{\textit{\textbf{FGWS}} \citep{mozes2021frequency}} \\
\hline
\textbf{Model} & \textbf{Dataset} & \textbf{Attack} & \textbf{F1-Score} & \textbf{Adv.} & \textbf{F1-Score} & \textbf{Adv.}\\
\textbf{} & \textbf{} & \textbf{} & \textbf{} & \textbf{Recall} & \textbf{} & \textbf{Recall}\\
\hline
DistilBERT & IMDb & PWWS & \textbf{92.1 $\pm$ 0.5} & 94.2 $\pm$ 1.1 & 89.5 & 82.7 \\
\hline
\hline
\cellcolor{Turquoise!30}LSTM & IMDb & PWWS & \textbf{84.1 $\pm$ 3.4} & 86.8 $\pm$ 8.5 & 80.0 & 69.6 \\
\hline
\cellcolor{Turquoise!30}CNN & IMDb & PWWS & 84.3 $\pm$ 3.1 & 90.0 $\pm$ 6.2 & \textbf{86.3} & 79.6 \\
\hline
\cellcolor{Turquoise!30}BERT & IMDb & PWWS & \textbf{92.4 $\pm$ 0.7} & 92.5 $\pm$ 1.8 & 89.8 & 82.7 \\
\hline
DistilBERT & \cellcolor{Turquoise!30}AG News & PWWS & \textbf{93.1 $\pm$ 0.6} & 96.1 $\pm$ 2.2 & 89.5 & 84.6 \\
\hline
DistilBERT & \cellcolor{Turquoise!30}RTMR & PWWS & 74.1 $\pm$ 3.1 & 85.1 $\pm$ 8.6 & \textbf{78.9} & 67.8 \\
\hline
DistilBERT & IMDb & \cellcolor{Turquoise!30}TextFooler & \textbf{94.2 $\pm$ 0.8} & 97.3 $\pm$ 0.9 & 86.0 & 77.6 \\
\hline
DistilBERT & IMDb & \cellcolor{Turquoise!30}IGA & \textbf{88.5 $\pm$ 0.9} & 95.5 $\pm$ 1.3 & 83.8 & 74.8 \\
\hline
DistilBERT & IMDb & \cellcolor{Turquoise!30}BAE & \textbf{88.0 $\pm$ 0.9} & 96.3 $\pm$ 1.0 & 65.6 & 50.2 \\
\hline
DistilBERT & \cellcolor{Turquoise!30}RTMR & \cellcolor{Turquoise!30}IGA & \textbf{70.4 $\pm$ 5.5} & 90.2 $\pm$ 6.9 & 68.1 & 55.2 \\
\hline
DistilBERT & \cellcolor{Turquoise!30}RTMR & \cellcolor{Turquoise!30} BAE & \textbf{68.5 $\pm$ 4.3} & 82.2 $\pm$ 9.0 & 29.4 & 18.5 \\
\hline
DistilBERT & \cellcolor{Turquoise!30}AG News & \cellcolor{Turquoise!30}BAE & \textbf{81.0 $\pm$ 4.3} & 95.4 $\pm$ 3.8 & 55.8 & 44.0 \\
\hline
\cellcolor{Turquoise!30}BERT & \cellcolor{Turquoise!30}YELP & PWWS & 89.4 $\pm$ 0.6 & 85.3 $\pm$ 1.7 & \textbf{91.2} & 85.6 \\
\hline
\cellcolor{Turquoise!30}BERT & \cellcolor{Turquoise!30}YELP & \cellcolor{Turquoise!30}TextFooler & \textbf{95.9 $\pm$ 0.3} & 97.5 $\pm$ 0.6 & 90.5 & 84.2 \\
\hline
\end{tabularx}
\caption{Performance results for detector trained on (DistilBERT, IMDb, PWWS).}
\label{tab:imdb_results}
\end{subtable}
\newline
\vspace*{0.5 cm}
\newline
\begin{subtable}{\textwidth}
\begin{tabularx}{\textwidth}{|X|X|X|X|X|X|X|}
\hline
\multicolumn{3}{|c|}{\textit{\textbf{Configuration}}} & \multicolumn{2}{c}{\textit{\textbf{WDR}} (Ours)} & \multicolumn{2}{|c|}{\textit{\textbf{FGWS}} \citep{mozes2021frequency}} \\
\hline
\textbf{Model} & \textbf{Dataset} & \textbf{Attack} & \textbf{F1-Score} & \textbf{Adv.} & \textbf{F1-Score} & \textbf{Adv.}\\
\textbf{} & \textbf{} & \textbf{} & \textbf{} & \textbf{Recall} & \textbf{} & \textbf{Recall}\\
\hline
DistilBERT & AG News & PWWS & \textbf{93.6 $\pm$ 1.5} & 94.8 $\pm$ 2.4 & 89.5 & 84.6 \\
\hline
\hline
\cellcolor{Turquoise!30}LSTM & AG News & PWWS & \textbf{94.0 $\pm$ 1.0} & 94.2 $\pm$ 2.2 & 88.9 & 84.9 \\
\hline
\cellcolor{Turquoise!30}CNN & AG News & PWWS & \textbf{91.1 $\pm$ 1.4} & 91.2 $\pm$ 2.6 & 90.6 & 87.6 \\
\hline
\cellcolor{Turquoise!30}BERT & AG News & PWWS & \textbf{92.5 $\pm$ 0.9} & 93.0  $\pm$ 1.8 & 88.7 & 83.2 \\
\hline
DistilBERT & \cellcolor{Turquoise!30}IMDB & PWWS &\textbf{91.4 $\pm$ 0.6} & 93.0 $\pm$ 1.9 & 89.5 & 82.7 \\
\hline
DistilBERT & \cellcolor{Turquoise!30}RTMR & PWWS & 75.8 $\pm$ 0.9 & 78.5 $\pm$ 4.8 & \textbf{78.9} & 67.8 \\
\hline
DistilBERT & AG News & \cellcolor{Turquoise!30}TextFooler & \textbf{95.7 $\pm$ 0.7} & 97.3 $\pm$ 1.2 & 87.0 & 79.4 \\
\hline
DistilBERT & AG News & \cellcolor{Turquoise!30}BAE & \textbf{86.4 $\pm$ 1.1} & 94.5 $\pm$ 1.8 & 55.8 & 44.0 \\
\hline
DistilBERT & AG News & \cellcolor{Turquoise!30}IGA & \textbf{86.7 $\pm$ 1.5} & 93.6 $\pm$ 2.1 & 68.6 & 58.3 \\
\hline
DistilBERT & \cellcolor{Turquoise!30}RTMR & \cellcolor{Turquoise!30}IGA & \textbf{73.7 $\pm$ 1.5} & 85.4 $\pm$ 5.2 & 68.1 & 55.2 \\
\hline
DistilBERT & \cellcolor{Turquoise!30}RTMR & \cellcolor{Turquoise!30} BAE & \textbf{71.0 $\pm$ 1.1} & 75.2 $\pm$ 6.0 & 29.4 & 18.5 \\
\hline
DistilBERT & \cellcolor{Turquoise!30}IMDB & \cellcolor{Turquoise!30}BAE & \textbf{88.1 $\pm$ 0.9} & 97.0 $\pm$ 1.0 & 65.6 & 55.2 \\
\hline
\cellcolor{Turquoise!30}BERT & \cellcolor{Turquoise!30}YELP & PWWS & 86.2 $\pm$ 1.4 & 77.2 $\pm$ 3.1 & \textbf{91.2} & 85.6 \\
\hline
\cellcolor{Turquoise!30}BERT & \cellcolor{Turquoise!30}YELP & \cellcolor{Turquoise!30}TextFooler & \textbf{95.4 $\pm$ 0.3} & 94.7 $\pm$ 0.9 & 90.5 & 84.2 \\
\hline
\end{tabularx}
\caption{Performance results for detector trained on (DistilBERT, AG News, PWWS).}
\label{tab:agnews_results}
\end{subtable}
\caption{Adversarial detection performance of our defense against the state of the art \textit{FGWS} under several setups. Results were obtained with a detector trained on two different configurations as indicated in the first row of each table. For all other rows, i.e. test configurations, differences w.r.t the training setup have been \textcolor{Turquoise}{highlighted}. To increase the results' statistical significance, we average the performance across 30 different data-splits of the training configuration. Additionally, we report the corresponding 95\% confidence intervals. Given the deterministic nature of \textit{FGWS}, different data-splits lead to the same performance and hence confidence intervarls are not reported as they are trivial ($\pm 0$).}
\end{table*}

As shown in Table \ref{tab:detector}, all architectures achieve competitive performance and none of them clearly appears superior to the others. We pick XGBoost \citep{chen2016xgboost} as it exhibits the best F1-score. The main hyperparameters utilized are $29$ gradient boosted trees with a maximum depth of $3$ and $0.34$ as learning rate. We utilize this detector architecture for all experiments in the following sections. 

\subsection{Detection Performance} \label{subsec:performance}

Tables \ref{tab:imdb_results} and \ref{tab:agnews_results} report the detection performance of our method in a variety of configurations. In each table, the first row represents the setting---i.e. combination of target model, dataset, and attack type---in which the detector was trained. The remaining rows, instead, are w.r.t. settings in which we tested the already trained detector without performing any kind of fine-tuning or retraining.

We utilize a balanced training set of size $3,000$ and $2,400$ samples respectively for the detectors trained on IMDb adversarial attacks (Table \ref{tab:imdb_results}) and on AG News attacks (Table \ref{tab:agnews_results}). All results are obtained using balanced test sets containing $500$ samples. The only exceptions are the configurations (DistilBERT, RTMR, IGA) 
and (DistilBERT, AG News, IGA) which used test sets of size $480$ and $446$ respectively due to data availability.


To the best of our knowledge, the FGWS method from \citet{mozes2021frequency} is the best detector available and was already proven to be better than DISP \citep{zhou2019learning} by its authors. Hence, we utlize FGWS as baseline for comparison in all configurations. Analogously to our method, FGWS is trained on the configuration in the first row of each table and then applied to all others. More in detail, we fine-tune its \emph{frequency substitution threshold} parameter $\delta$ \citep{mozes2021frequency} until achieving a best fit value of $\delta = 0.9$ in both training settings.

From what can be seen in both tables, the proposed method consistently shows very competitive results in terms of F1-score and outperforms the baseline in $22$ configurations out of $28$ (worse in $5$) and is on average better by $8.96$ percentage points. At the same time, our methods exhibits a very high adversarial recall, showing a strong capability at identifying attacks and thus producing a small amount of false negatives.

\paragraph{Generalization to different target models:} Starting from the training configurations, we vary the \emph{target model} while maintaining the other components fixed (rows 2-4 of each table). Here, the detector achieves state-of-the-art results in all test settings, occasionally dropping below the $90\%$ F1-score on a few simpler models like LSTM and CNN while not exhibiting any decay on more complex models like BERT. 

\paragraph{Generalization to different datasets:} Analogous to the previous point, we systematically substitute the \emph{dataset} component for evaluation (rows 5-6 of each table). We notice a substantial decay in F1-score when testing with RTMR (74.1 - 75.8\%) since samples are short and, therefore, may contain few words which are very relevant for the prediction, just like adversarial replacements. Nevertheless, removing adversarial words still result in a change of prediction to the original class thereby preserving high adversarial recall."

\paragraph{Generalization to different attacks:} Results highlight a good reaction to all other text attacks (rows 7-9 of each table) and even experiences a considerable boost in performance against TextFooler. In contrast, the baseline \textit{FGWS} significantly suffers against more complex attacks such as BAE, which generates context-aware perturbation.

Besides testing generalization properties via systematically varying one configuration component at the time, we also test on a few settings presenting changes in multiple ones (rows 10-14 of each table). Also in these settings, the proposed method maintains a very competitive performance, with noticeable drops only on the RTMR dataset.

\subsection{Tuning the Decision Boundary} \label{subsec:boundary}

Depending on the application in which the detector is used to monitor the model and detect malicious input manipulations, different performance metrics can be taken into account to determine whether it is safe to deploy the model. For instance, in a very safety-critical application where successful attacks lead to harmful consequences, \emph{adversarial recall} becomes considerably more relevant as a metric than the F1-score. 

\begin{figure}[!h]
    \centering
    \includegraphics[width=0.99\linewidth]{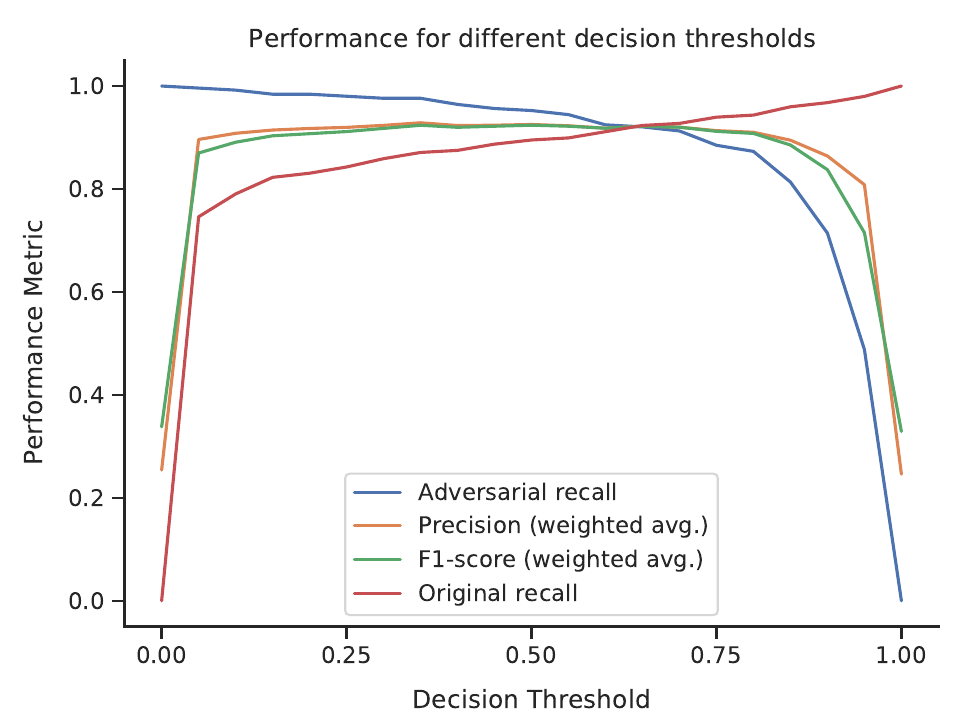}
    \caption{Performance metrics w.r.t. different decision thresholds for our XGBoost classifier on the configuration (IMDb, DistilBERT, PWWS). Input sentences are classified as adversarial when their probability is higher than the decision threshold.}
    \label{fig:metrics}
\end{figure}

We examine how relevant metrics change in response to different choices for the discrimination threshold. Please note that a lower value corresponds to more caution, i.e. we are more likely to output that a certain input is adversarial.

\begin{table}[h]
\centering
\begin{tabular}{|l|l|l|l|l|}
\hline
\textbf{DT} & \textbf{Precision} & \textbf{F1} & \textbf{Adv.} & \textbf{Orig.}\\
\textbf{} & \textbf{} & \textbf{} & \textbf{Recall} & \textbf{Recall}\\
\hline
0.5 & 92.5 & 92.4 & 95.2 & 89.5 \\
\hline
\hline
0.4 & 92.3 & 92.0 & 96.4 & 87.5 \\
\hline
0.3 & 92.4 & 91.8 & 97.6 & 85.9 \\
\hline
0.15 & 91.5 & 90.3 & \textbf{98.4} & 82.3 \\
\hline
\end{tabular}
\caption{Performance comparison using different \emph{Decision Thresholds} (DT) for our XGBoost classifier on the configuration (IMDb, DistilBERT, PWWS). The used default value is 0.5.}
\label{tab:threshold}
\end{table}

Figure \ref{fig:metrics} and Table \ref{tab:threshold} show performance results w.r.t. different threshold choices. We notice that decreasing its value from 0.5 to 0.15 can increase the adversarial recall to over $98\%$ at a small cost in terms of precision and F1-score ($< 2$ percentage points). Applications where missing attacks---i.e. false negatives---have disastrous consequences could take advantage of this property and consider lowering the decision boundary. This is particularly true if attacks are expected with a low frequency and an increase in false positive incurs only minor costs.

\section{Discussion and Qualitative Results} \label{sec:discussion}

Section \ref{sec:results} discussed quantitative results and emphasized the competitive performance that the proposed approach achieves. Here, instead, we focus on the qualitative aspects of our research findings. For instance, we try to understand \textit{why} our pipeline works while also discussing challenges, limitations, ethical concerns, and future work.

\subsection{Understanding the Adversarial Detector} \label{subsec:shap}

The proposed pipeline consists of a machine learning classifier---e.g. XGBoost---fed with the model's WDR scores. The intuition behind the approach is that words replaced by adversarial attacks play a big role in altering the target model's decision. Despite the competitive detection performance, the detector is itself a learning algorithm and we cannot determine with certainty what patterns it can identify. 

To validate our original hypothesis, we apply a popular explainability technique---SHAP \citep{lundberg2017unified}---to our detector. This allows us to summarize the effect of each feature at the dataset level. We use the official implementation\footnote{https://github.com/slundberg/shap, released under MIT License} to estimate the importance of each WDR and use a \emph{beeswarm plot} to visualize the results.

\begin{figure}[h]
    \centering
    \includegraphics[width=0.98\linewidth]{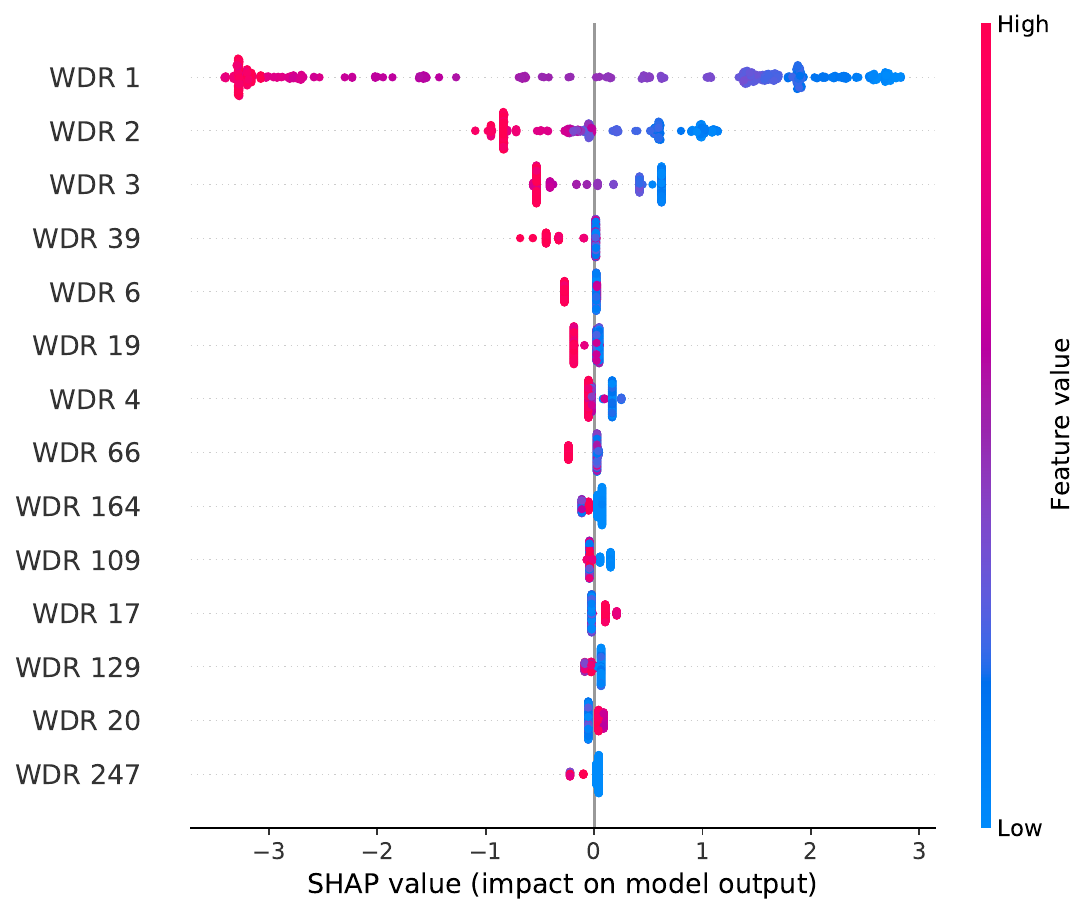}
    \caption{WDR scores with the highest impact (SHAP value) on the detector's prediction. Please recall that the WDR scores are sorted by magnitude. For instance, WDR 1 is the first and largest WDR score.}
    \label{fig:shap1}
\end{figure}

Figure \ref{fig:shap1} shows that values in the first positions---i.e. 1, 2, and 3---of the input sequence are those influencing the adversarial detector the most. Since in our pipeline WDR scores are sorted based on their magnitude, this means that the largest WDR of each prediction are the most relevant for the detector. This is consistent with our hypothesis that replaced words substantially change output logits and thus measuring their variation is effective for detecting input manipulations. As expected, negative values for the WDR correspond to a higher likelihood of the input being adversarial. 

We also notice that features after the first three do not appear in the naturally expected order. We believe this is the case as for most sentences it is sufficient to replace two-three words to generate an adversarial sample. Hence, in most cases, only a few WDR scores carry important signals for detection.

\subsection{Challenges and Limitations} \label{subsec:limitations}

While WDR scores contain rich patterns to identify manipulated samples, they are also relatively expensive to compute. Indeed, we need to run the model once for each feature---i.e. each word---in the input text. While this did not represent a limitation for our use-cases and experiments, we acknowledge that it could result in drawbacks when input texts are particularly long. 

Our method is specifically designed against word-level attacks and it does not cover character-level ones. However, the intuition seems to some extent applicable also to sentences with typos and similar artifacts as the words containing them will play a big role for the prediction. This, like in the word-level case, needs to happen in order for the perturbations to result in a successful adversarial text attack and change the target model's prediction


\subsection{Ethical Perspective and Future Work}

Detecting---or in general defending against---adversarial attacks is a fundamental pillar to deploy machine learning models ethically and safely. However, while defense strategies increase model robustness, they can also inspire and stimulate new and improved attack techniques. An example of this phenomenon is BAE \citep{garg2020bae}, which leverages architectures more resilient to attacks such as BERT to craft highly-effective contextual attacks. Analogously, defense approaches like ours could lead to new attacks that do not rely on a few words to substantially affect output logits.

Based on our current findings, we identify a few profitable directions for future research. \textbf{(1)} First of all, the usage of logits-based metrics such as the WDR appears to be very promising for detecting adversarial inputs. We believe that a broader exploration and comparison of other metrics previously used in computer vision could lead to further improvements. \textbf{(2)} We encourage future researchers to draw inspiration from this work and also test their defenses in settings that involve mismatched attacks, datasets, and target models. At the same time, we set as a priority for our future work to also evaluate the efficacy of adversarial detection methods on adaptive attacks \citep{tramer2020adaptive, athalye2018synthesizing}. \textbf{(3)} This work proves the efficacy of WDR in a variety of settings, which include a few different datasets and tasks. However, it would be beneficial for current research to understand how these techniques would apply to high-stakes NLP applications such as hate speech detection \citep{mosca2021understanding, wich2021explainable}.

\section{Conclusion} \label{sec:conclusion}

Adversarial text attacks are a major obstacle to the safe deployment of NLP models in high-stakes applications. However, although manipulated and original samples appear indistinguishable, interpreting the model's reaction can uncover helpful signals for adversarial detection.

Our work utilizes logits of original and adversarial samples to train a simple machine learning detector. WDR scores are an intuitive measure of word relevance and are effective for detecting text components having a suspiciously high impact on the output. The detector does not make any assumption on the classifier targeted by the attacks and can be thus considered model-agnostic.

The proposed approach achieves very promising results, considerably outperforming the previous state-of-the-art in word-level adversarial detection. Experimental results also show the detector to possess remarkable generalization capabilities across different target models, datasets, and text attacks without needing to retrain. These include transformer architectures such as BERT and well-established attacks such as PWWS, genetic algorithms, and context-aware perturbations. 

We believe our work sets a strong baseline on which future research can build to develop better defense strategies and thus promoting the safe deployment of NLP models in practice. We release our code to the public to facilitate further research and development \footnote{Public repository: https://github.com/javirandor/wdr}.

\bibliography{main}
\bibliographystyle{acl_natbib}

\end{document}